\documentclass[conference]{IEEEtran}
\IEEEoverridecommandlockouts
\usepackage{cite}
\usepackage{amsmath,amssymb,amsfonts}
\usepackage[utf8]{inputenc}
\usepackage[T1]{fontenc}
\usepackage{algorithm}
\usepackage{algorithmic}
\usepackage{graphicx}
\usepackage{textcomp}
\usepackage{xcolor}
\usepackage{varwidth}
\usepackage{multirow}
\usepackage{hyperref}
\def\BibTeX{{\rm B\kern-.05em{\sc i\kern-.025em b}\kern-.08em
    T\kern-.1667em\lower.7ex\hbox{E}\kern-.125emX}}

\begin{document}

\title{Classification Based on Association Rules Algorithm for Breast Cancer\\
}

\author{\IEEEauthorblockN{Ali Alsalama}
\IEEEauthorblockA{\textit{Department of Computer Science} \\
\textit{University of Sharjah}\\
Sharjah, UAE \\
a.slama1@hotmail.com}
\and
\IEEEauthorblockN{Ahmed Kubba}
\IEEEauthorblockA{\textit{Department of Computer Science} \\
\textit{University of Sharjah}\\
Sharjah, UAE \\
ahmedkubba@outlook.com}
\and
\IEEEauthorblockN{Ghaith Jamjoum}
\IEEEauthorblockA{\textit{Department of Computer Science} \\
University of Sharjah\\
Sharjah, UAE \\
U23102953@sharjah.ac.ae}
\and
\IEEEauthorblockN{Zaher Al Aghbari}
\IEEEauthorblockA{\textit{Department of Computer Science} \\
University of Sharjah\\
Sharjah, UAE \\
zaher@sharjah.ac.ae}
}

\maketitle

\begin{abstract}
Breast cancer is a significant contributor to female mortality across the world, displaying one of the highest occurrence rates among the various cancer types. In response to the need for early breast cancer detection, researchers have increasingly turned to association rule-based classification as a favored method. Association Rule mining is a data mining approach which offers the benefit of yielding results that are readily understandable for medical professionals. This paper introduces a novel association rule-based data mining technique for breast cancer classification based on a weighted classification approach. This implementation employs three core algorithms: Rule Generation, Rule Pruning, and Rule Prediction. Rule Generation identifies frequent itemsets and creates association rules. Rule Pruning eliminates rules using specific criteria and separates them into major and minor groups based on their influence on training data. Rule Prediction applies the pruned rules to classify test data. The final prediction algorithm was tested on several testing samples to show the feasibility and performance of the approach.
\end{abstract}


\section{Introduction}
Breast cancer poses a significant health challenge for women worldwide, displaying an annual increase at an unpredictable rate. The danger surrounding this disease has encouraged researchers to adopt novel techniques to address the issue. Among these methods are data mining approaches, such as association classification (AC). [1] Breast cancer stands as one of the most prevalent human neoplasms, constituting approximately a quarter of all female cancers. Moreover, it ranks as a leading cause of female mortality in many countries, necessitating intense efforts to mitigate the impact of this chronic disease. [2]

    Data mining serves as an important step in Knowledge Discovery from Databases (KDD), encompassing various approaches for diverse objectives. [3] Classification approaches work by predicting the class of an instance after training on a sizable dataset. On the other hand, association approaches uncover the relationships between different items in a dataset to unveil unknown patterns and generate new rules [20]. In order to improve the performance of the classification approach, which is crucial in fields like breast cancer, researchers have recently integrated association rule techniques into classification processes, introducing a novel approach known as the Association Classification technique. [3]

The AC approach integrates association rule mining into the classification process to uncover unknown patterns which can enhance classification accuracy. Accuracy is crucial in the decisive processes across various implementations and domains. The AC technique is designed to establish correlations between items and classes. For instance, consider the rule R: A, B → Class1, means that if attributes A and B co-occur in an instance, then it is considered that the instance is classified in Class1. [4]

\subsection{Association Classification}

Typically, most AC algorithms utilize two key measures, namely support and confidence, for generating association rules, without specific consideration for the application domain.  Moreover, in order to identify the significance of attributes within a given dataset, they do not have an appropriate preference technique.  AC is considered a fusion of association rules and classification methods. Association rules aim to identify correlations between attributes, while classification techniques predict class labels. [5]

In the AC problem, if a rule states At1 → C1, it can be inferred that C1 is a class attribute. The training dataset T is composed of k distinct attributes, (A, B, ..., K), with C representing a set of classes. Attributes may be classified as either categorical or continuous. Each training object in dataset T is represented by a vector consisting of attributes and an associated class label. The actual occurrence (actoccr) of the itemset rule r in T is the number of rows in T matching the itemsets defined in r. The support count (suppcount) of itemset rule r is the number of rows in T that match r's itemsets and belong to class Ci for r.

The formula for calculating \textbf{support }is given by:
\begin{equation}
\textit{Support = suppcount(r) / T}
\end{equation}
where suppcount(r) represents the number of transactions that contain the rule (r), and T is the total number of transactions in the dataset.

On the other hand, the formula for calculating confidence is:
\begin{equation}
Confidence = suppcount(r) / actoccr(r)
\end{equation}
where actoccr(r) represents the number of transactions where the antecedent of the rule (r) occurs.

The equation for \textbf{support count }is given as :
\begin{equation}
\textit{SupportCount = r }\cup \textit{ci}
\end{equation}
 If itemset rule exceeds the minsupp threshold, the following condition must be fulfilled, where |T| represents the number of instances in T:
\begin{equation}
\textit{(suppcount(r) / |T|)} \geq \textit{minsupp}
\end{equation}
Similarly, to determine if an itemset rule passes the \textbf{minconf }threshold, the following condition is considered, where actoccr(r) represents the actual occurrence of the itemset rule:
\begin{equation}
\textit{(suppcount(r) / actoccr(r))} \geq \textit{minconf}
\end{equation}
 A \textbf{frequent itemet rule} shall be defined as any itemet rule that meets the minsupp threshold. An actual class association rule is expressed in the following form, where the antecedent is an itemset, and the consequent is a class:

\begin{equation}
(A, v_{i1}) \wedge (B, v_{i2}) \wedge \ldots \wedge (K, v_{ik}) \rightarrow C_j
\end{equation}

\subsection{Breast Cancer}

The biggest concern in medicine is the early detection of diseases. Early diagnosis, particularly in the case of diseases like cancer, has the potential to significantly reduce the mortality rate among patients. Cancers are broadly categorized into two types, with the first type being benign cancer, and the second type being malignant cancer.

Breast cancer stands out as one of the most dangerous forms of cancer affecting women globally, prevailing as the primary cancer in females worldwide and affecting about 10\% of women across the world at some time in their lives. It ranks as the second leading cause of deaths related to cancer in women. Malignant tumours in breast cancer develop as cells in the breast tissue undergo uncontrolled growth and proliferation without proper regulation of cell death. While the precise causes of the majority of breast cancers remain unknown, certain risk factors are identified. [6]

Breast cancer research typically adopts a biological or clinical approach, frequently utilizing statistical patient data. Predicting the recurrence of the disease is a complex and intriguing task that forms the basis for developing data mining applications. [19] Improving the accuracy of breast cancer recurrence prediction not only alleviates patient anxiety but also plays a pivotal role in the pathological assessment. Utilizing data mining techniques, the recurrence of breast cancer can be predicted by analysing various indicators derived from statistical results. [7]

\subsection{Weighted Association Rule Mining}

In the conventional realm of mining rules, the significance of a rule is typically assessed by the frequency of its occurrences in a database. Traditional methods rely on measures like support and confidence to uncover frequent itemsets, assuming equal importance for all items. Conversely, social scientists and business researchers take a different stance. They gauge the importance of a rule not only by quantitative aspects, such as the item appearances in a database, but also on qualitative factors, like human interpretation. Introducing weights allows for this representation of item knowledge within a dataset.

Wang et al. [19] introduced the Weighted Association Rule Mining technique (WARM), which assigns weights to database items, which reflect an item's significance in that database.  In this approach, the data sets undergo a weight generation process based on domain knowledge and are assigned weights to individual items in the database. These weights are then used as input for WARM algorithms. In order to derive a set of compelling rules, this algorithm incorporates weights into the items and uses weighted support. The aim is to focus the mining process on significant associations, taking into account items of particular importance.

Most data items lack predefined weights, leaving them open to adjustments by domain experts. This flexibility is intriguing because experts can assign varied weights, thereby generating diverse rules. In certain scenarios, determining precise weights for all items using domain knowledge might be unfeasible, particularly with datasets containing many items. In such instances, a semi-automated or automated approach becomes more practical. Taking these factors into account, J. Alwidian [8] expanded upon CBA (Classification Based on Associations) and introduced a new weighting algorithm termed WCBA (Weighted CBA).

\subsection{Contributions}

The contributions of this paper are as follows:

 1. Many papers in the current literature split the rules into strong and weak rules based on different methods, whereas this implementation uses a custom classification for rules, splitting the rules into `major' and `minor' rules based on their coverage of the training dataset instances as described in the previous section, defining rules that do not match a single training dataset instance being categorized as `minor' rules.

 2. This implementation uses a custom approach for determining a default class (malignant or benign) which is based on the remaining non-discarded training dataset instances after the partial pruning process, choosing the class with the highest number of remaining instances as the default class. During the pruning process, conflicting malignant and benign rules are removed.

 3. This implementation removed training data instances linked to the major rules during the partial pruning process, which reduces rule redundancy and hence makes the algorithm more resource efficient.

 \subsection{Paper Structure}

The paper is organized as follows: In section 2 the related work is discussed and analyzed, including the limitations of the existing literature. In section 3, the paper's implementation is explained based on three main defined algorithms. In section 4, the dataset used in the experiment is discussed and data visualization is presented to better understand the distribution of the classes and features in the dataset.

\section{Related Work}

Alwidan et al. [8] introduced the Weighted Classification Based on Association Rules (WCBA) algorithm, which incorporates attribute prioritization through expert-assigned weights. The algorithm employs dual classification to identify key attributes, minimizing errors, and integrates a statistical measure to improve rule generation.

Pala et al. [9] used the apriori algorithm to extract class-specific rules from medical data using the UC-Irvine "Breast Cancer" dataset with 9 attributes and 1 class attribute from 286 patients. Using the Weka data mining program, the study generated valuable rules for preventing BC recurrence, such as the absence of a lymph node capsule ensuring a 100\% non-recurrence rate.

Umesh et al. [10] applied association rule mining to the SEER dataset, cleaning it with the Expectation Maximization (EM) method. The study utilized three sets of 547 records for training and reported a minimal classification error rate of 0.0724. Importantly, the generated rules successfully classify patients into two distinct classes: recurrence and non-recurrence. Shrivastav et al. [11] presented an overview of data mining techniques in the context of BC data analysis, emphasizing the use of decision trees in an experimental setup. Like Pala et al. [9], they employed the Weka data mining software, using a dataset with 10 attributes and 699 instances from the UCI machine learning data repository. This study underscores the practicality and accessibility of Weka for various data mining functionalities.

Khalilian et al.[12] concentrate on early breast cancer (BC) detection using the FP-growth algorithm to generate association rules for classifying breast masses as benign or malignant. Their approach combines association rule mining and classification techniques, emphasizing the practical use of the FP-Growth algorithm in medical data mining. The study highlights its potential significance for accurate BC diagnosis, particularly in early detection and healthcare decision-making.

Li et al.[13] address BC prevention and control through a three-level approach, introducing association rule-based algorithms to detect risk factors. The study emphasizes the significance of these rules in assessing BC risk and highlights the application of machine learning for risk assessment, demonstrating its superiority over traditional models. Their proposed system, leveraging information technology, aims to facilitate early detection and control by identifying established and newly discovered risk factors.

Karabatak et al.[14] present an automatic diagnosis system for detecting BC based on association rules and neural networks. They use association rules to reduce the dimensionality of the BC database, which counteracts the effect of large input features on classification accuracy [21], and they employ neural networks for classification. The performance comparison for BC detection shows varying correct classification rates, with the combination of association rules (AR1 and AR2) and neural networks achieving higher accuracy compared to using neural networks alone. Palaniappan [15] builds upon the work of Karabatak et al. [14] by introducing a fourth combination of association rules (AR1 and AR2) and neural networks, resulting in a superior 98.4\% correct rate for BC detection. This suggests the effectiveness of combining association rule mining techniques with neural networks for improved accuracy in BC diagnosis.

Malpani et al. [16] introduced two association rules mining approaches utilizing pre-processed data and the Apriori algorithm. In the first approach, they used WEKA to generate association rules (ARs) between a single gene and a set of transcriptional factors (TF). The second approach involved Excel filtering based on a p-value threshold to generate ARs between sets of genes and TFs. Through four experiments, various ARs were obtained, showing confidence levels and average performance rates ranging from 60\% to 100\%

Ruijuan Hu [17] identified deficiencies in the Apriori algorithm and proposed an improved version to reduce the size of candidate sets. They conducted data mining on breast cancer (BC) patient data provided by the Health Information System (HIS) using SQL Server 2005 Analysis Services. Four association rules were derived, with one having a confidence of 61.5\%, and three with a 100\% confidence. The results suggest that patients with smaller tumor sizes and those receiving radiotherapy have fewer possibilities of recurrence, as indicated by the association rules.

In the BC datasets examined in the literature, there is a recognized need for additional relevant attributes and improved approaches for handling missing values to enhance the prediction of BC recurrence. Bayesian networks have proven widely effective in medical prediction, particularly in BC prognosis and diagnosis. The suggestion to integrate Association Rules with Bayesian Networks is proposed as a promising avenue for future research, akin to the successful combination of association rules, and neural networks as demonstrated in the study by Karabatak et al. [14]. The variability in correct classification rates observed in their study suggests the potential for achieving improved accuracy by combining different techniques, such as Bayesian Networks, with diverse data mining approaches in BC detection research. Overall, these studies demonstrate different approaches to breast cancer detection, with a common emphasis on leveraging association rule mining and machine learning techniques. Each approach offers unique contributions, highlighting the potential of combining these methods for more accurate and effective BC diagnosis and risk assessment.

\section{AR for Breast Cancer Classification}

The implementation, which is an association rules classification approach, is based on three main algorithms responsible for rule generation, pruning, and prediction.

\subsection{Rule Generation}

\begin{algorithm}
\caption{Algorithm 1 - Rule Generation algorithm pseudocode}
\label{alg:rule_generator}
\begin{algorithmic}[1]
\REQUIRE mini\_Support, df, itemset\_df, weights
\ENSURE itemset\_df, frequent\_itemsets\_benign\_count, frequent\_itemsets\_malignant\_count, candidate\_benign\_itemsets, candidate\_malignant\_itemsets
\STATE $k \leftarrow 1$ \label{alg:line:init_k}
\WHILE{itemsets exist} \label{alg:line:while}
    \STATE Initialize itemset\_benign \label{alg:line:init_benign}
    \FOR{each transaction in the dataset} \label{alg:line:for_transaction}
        \FOR{each itemset in itemsets} \label{alg:line:for_itemset}
            \IF{itemset is present in the transaction AND the transaction indicates benign} \label{alg:line:if_benign}
                \STATE Increment count for itemset in itemset\_benign \label{alg:line:increment_benign}
                \STATE Increment count for itemset in candidate\_benign\_itemsets \label{alg:line:increment_candidate}
            \ENDIF
        \ENDFOR
    \ENDFOR
    \STATE Store counts in a list for each itemset and its count in itemset\_benign \label{alg:line:store_counts}
    \FOR{each itemset and its count in itemset\_benign} \label{alg:line:for_count}
        \STATE Calculate weighted support \label{alg:line:calculate_support}
    \ENDFOR
    \STATE Filter frequent itemsets based on minimum support \label{alg:line:filter}
    \STATE Update frequent itemset counts \label{alg:line:update_counts}
    \STATE Increment k \label{alg:line:increment_k}
    \STATE Generate new itemsets from frequent itemsets \label{alg:line:generate_new}
\ENDWHILE
\RETURN itemset\_df, frequent\_itemsets\_benign\_count, frequent\_itemsets\_malignant\_count, candidate\_benign\_itemsets, candidate\_malignant\_itemsets \label{alg:line:return}
\end{algorithmic}
\end{algorithm}

The Rule Generation algorithm follows a two-step process, involving the identification of frequent itemsets—sets of items occurring together with high frequency—and the generation of association rules based on these itemsets. By iteratively applying a join and prune strategy, the algorithm efficiently extracts patterns in data, revealing relationships between items. The generated association rules, in the form of "if-then" statements, separately for benign and malignant classes, also utilize a weighted support threshold.

The weighted support is calculated based on a predefined set of weights assigned to specific columns in the dataset, which were based on the weighted classification based on association rules (WCBA) approach, which were acquired through domain experts from King Hussein Cancer Center in Amman. [8] The function also performs the necessary data pre-processing on the breast cancer diagnosis dataset by dropping the unnecessary columns, converting the dataset into a list of lists, and applying the set of pre-defined weights to specific attribute columns in the dataset.

\subsection{Rule Pruning}

The Rule Pruning algorithm is a custom rule pruning implementation that takes the results of the rule generation process (frequent itemsets) for two classes ("benign" and "malignant") and prunes the rules based on minimum confidence and harmonic mean values. The function first prunes the rules based on a defined minimum confidence threshold, and then it calculates the harmonic mean for each rule and sorts the rules based on this value. The algorithm then checks for conflicts between benign and malignant rules, deletes the conflicting rules, and performs partial pruning, which splits the rules into `major' and `minor' rules based on their presence in the training dataset instances.

During the partial pruning process, a rule is included into the classifier if it encompasses at least one training dataset instance, and once a rule is integrated into the classifier, all training instances linked to it are eliminated. This iterative process continues until either all training dataset instances have been discarded, or all the rules have been tested.

In case there are rules which did not match a single training instance, these rules are categorized as `minor' rules. On the other hand, if there are remaining training dataset instances after the partial pruning process has finished, the algorithm counts the malignant and benign remaining instances and chooses the class with the highest number of training instances left as the default class.

\subsection{Rule Prediction}

The Rule Prediction algorithm utilizes the set of rules generated and pruned by previous algorithms. The function takes the major and minor rules for benign and malignant classes, along with the test instance, and identifies rules from each class that match the features of the test instance. Then it calculates the harmonic mean average for the matched rules. The function then compares the harmonic mean averages for both benign and malignant rules and classifies the test instance accordingly, considering a default class if needed.

\section{Experiments}

This section details the training and testing dataset, the experimental setup, and the experimental results of the breast cancer classification algorithm.

\subsection{Dataset}

\begin{figure}
    \centering
    \includegraphics[width=1\linewidth]{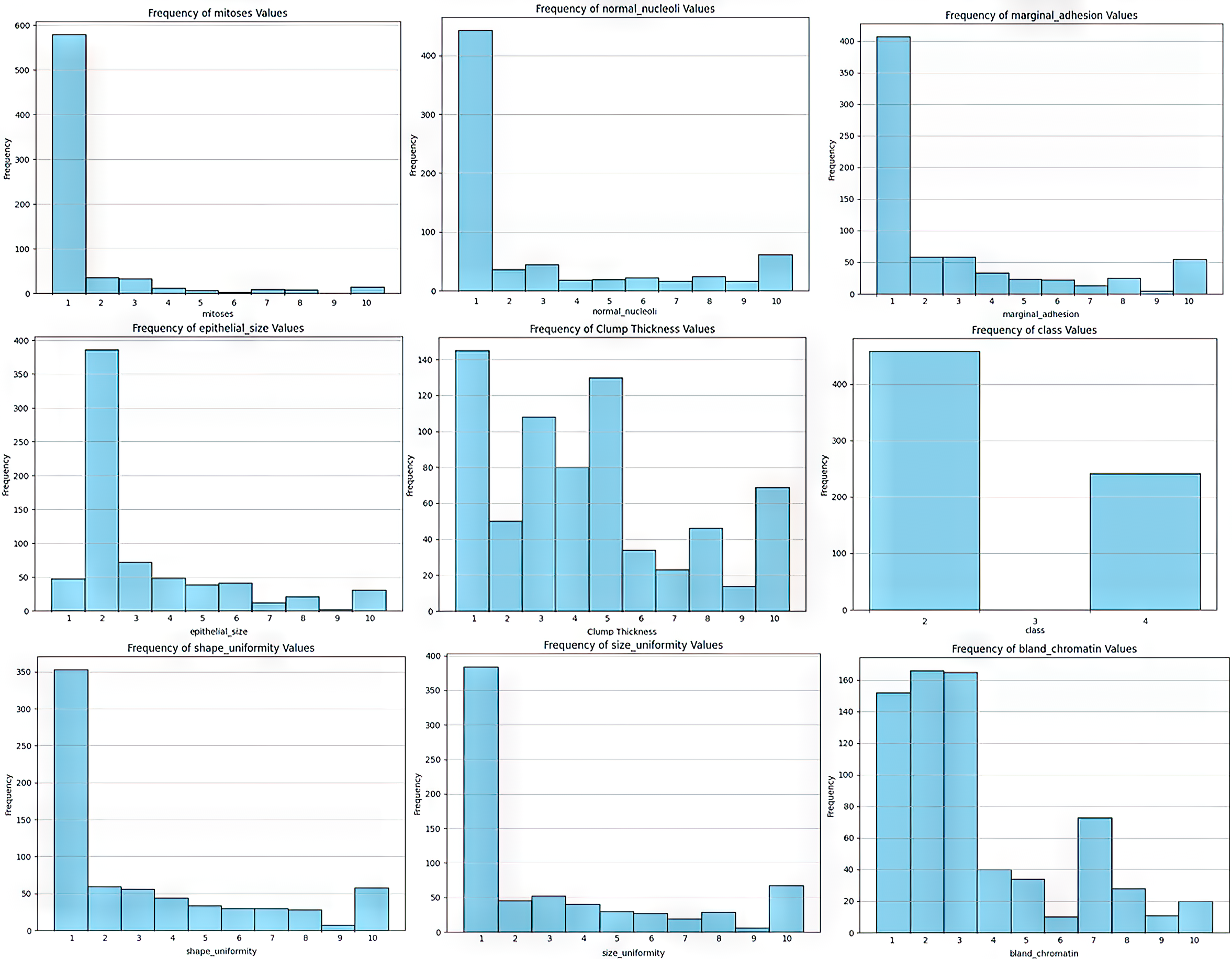}
    \caption{Visualization of the distribution of the breast cancer diagnosis dataset's features. The x-axis represents the expression level for each attribute, whereas the y-axis represents the frequency or count of that attribute at a certain expression level in the dataset as a whole}
    \label{fig:enter-label}
\end{figure}

The dataset used in this paper's experiment is the Breast Cancer Wisconsin (Diagnostic) dataset from the UCI repository [18], which contains 10 attributes and 699 instances. The dataset contains attributes related to breast cancer cells, each assigned a value on a scale from 1 to 10. The visualized distribution of the dataset can be observed in Fig.~\ref{fig:enter-label}

Clump Thickness indicates the thickness of cell clusters, while Cell Size Uniformity and Cell Shape Uniformity measure the consistency of cell sizes and shapes, respectively. Marginal Adhesion reflects how well cells stick together, and Single Epithelial Cell Size quantifies the size of individual epithelial cells. Bare Nuclei denotes the presence of nuclei in cells without surrounding cytoplasm, and Bland Chromatin describes the appearance of chromatin in the cell nucleus. Normal Nucleoli indicates the appearance of normal cell nuclei, and Mitoses represents the number of mitotic figures observed, reflecting cell division. Finally, the attribute "Class benign and malignant" serves as a classification label, distinguishing between benign (non-cancerous) and malignant (cancerous) cells. These attributes are pivotal in the context of breast cancer diagnosis, where machine learning models leverage them to predict the likelihood of malignancy based on cell characteristics. A visualization of the distribution of the dataset's attributes can be observed in Fig. 1.

The breast cancer dataset suffers from a clear imbalance in that the benign class samples far outnumber the malignant samples which can potentially cause some issues in generalization and the rules generating process. During the implementation of this paper, we sought the advice of domain experts in the medical field and they have told us that data augmentation when it comes to medical data is not acceptable or very difficult, hence we could not solve the dataset imbalance using that approach.

After using the dataset to generate the association rules and pruning them, selected test samples were taken from the breast cancer dataset and used to evaluate the implementation's prediction algorithm in order to ensure that it works as expected and correctly predicts the test sample's class label, whether malignant or benign. The prediction algorithm correctly predicted the class label for the given test samples, indicating success in the tests.

\subsection{Experimental Setup}

The training and testing were performed on a virtual machine running a Linux Ubuntu operating system, with 4 GB of RAM and an AMD Ryzen quadcore processor.

 The testing was conducted on a categorical basis, classifying several attributes into different categories from category 1 to 3. For evaluation, the accuracy and F1 score metrics were used to evaluate the algorithm's performance and compare it to other works.

 Category 1 represents one to three attributes as input, whereas Category 2 represents four to six attributes and Category 3 represents seven to nine input attributes. The tests were performed on different support values (10\%, 20\%, and 30\%), with a fixed confidence value of 50\%.

 First, 140 samples were used for the testing data, and any missing values in the data samples were ignored by being replaced with zeroes. The values of the nine attributes were discretized from 1 to 10, and a function was used which adds zeroes to one or two attributes to a random sample in the testing data within category 3 data samples such that the data's attributes range from 1 to 3. For category 2, the function adds random zeroes to the testing samples so that the attributes range from 4 to 6 attributes. Finally, for Category 1 the same process was applied to make the samples have attributes that range from 1 to 3 attributes.

\subsection{Experimental Results}

The final results of the tests conducted to evaluate the algorithm's performance can be observed in Table~\ref{tab:my_table}.

\begin{table}
\centering
\caption{Testing results and evaluation metrics of the algorithm.}
\begin{tabular}{|l|l|l|l|}
\hline
Category & Support & Accuracy & F1 Score \\
\hline
\multirow{3}{*}{Category 1} & 10\% & 0.908 & 0.851 \\ \cline{2-4}
 & 20\% & 0.950 & 0.923 \\ \cline{2-4}
 & 30\% & 0.865 & 0.759 \\ \hline
\multirow{3}{*}{Category 2} & 10\% & 0.972 & 0.959 \\ \cline{2-4}
 & 20\% & 0.943 & 0.915 \\ \cline{2-4}
 & 30\% & 0.908 & 0.847 \\ \hline
\multirow{3}{*}{Category 3} & 10\% & 0.936 & 0.905 \\ \cline{2-4}
 & 20\% & 0.950 & 0.923 \\ \cline{2-4}
 & 30\% & 0.922 & 0.874 \\ \hline
\end{tabular}
\label{tab:my_table}
\end{table}
As the support increase in Category 1 from 10\% to 20\%, the accuracy and F1 both improve. However, further increasing the support to 30\% decreases the performance metrics below their original values, which indicates that the ideal support value is around 20\% for Category 1. The same observation can be made for Category 3, in which a support value of 20\% results in the highest accuracy and F1 score compared to the other support value. In Category 2, increasing the support value consistently results in lower performance in terms of accuracy and F1 score, suggesting that a support value of around 10\% is ideal for Category 2.

In general, it can be observed that as the support percentage increases, the performance metrics tend to decrease or remain stable. However, this trend is not necessarily consistent across all categories and metrics, which makes it essential to consider the specific characteristics of each category and the effects of changing the support percentage on the performance metrics.

\section*{Conclusion}

Breast cancer is a serious health issue that is more widespread today relative to other forms of cancer in women. Due to its dangerous symptoms and annual increase among the world population, there is a great amount of interest in researching better techniques and approaches for the detection and classification of breast cancer and its stages. This paper presents a novel association classification-based implementation for breast cancer classification, with a custom approach for choosing the default class and dividing the rules into major and minor. The prediction algorithm correctly predicted the test sample's class label, indicating that this implementation was successful in its purpose to create an Association Rules Classifier for breast cancer by building on the weighted classification method [8] and introducing a combination of new novel methods and unique techniques.

\end{document}